\pdfoutput=1 
\documentclass[11pt,a4paper]{article}

\usepackage[hyperref]{emnlp2020}

\usepackage[utf8]{inputenc}
\usepackage{CJKutf8}

\usepackage{times}
\usepackage{latexsym}

\usepackage{microtype}
\usepackage{amsmath}

\usepackage{graphicx}
\usepackage{tikz}
\usepackage{pgfplots}
\pgfplotsset{compat=1.16}

\usepackage{booktabs}
\usepackage{array}

\aclfinalcopy

\title{Look It Up: Bilingual Dictionaries Improve Neural Machine Translation}

\author{Xing Jie Zhong \\
University of Notre Dame \\
  \texttt{xzhong3@nd.edu} \\\And
  David Chiang \\
University of Notre Dame \\
  \texttt{dchiang@nd.edu} \\}

\date{}

\newcommand{\unk}{\texttt{UNK}}

\newcommand{\zh}[1]{\begin{CJK*}{UTF8}{gbsn}#1\end{CJK*}}

\usepackage{newtxmath}

\begin{document}

\maketitle
\begin{abstract}
Despite advances in neural machine translation (NMT) quality, rare words continue to be problematic. For humans, the solution to the rare-word problem has long been dictionaries, but dictionaries cannot be straightforwardly incorporated into NMT. In this paper, we describe a new method for ``attaching'' dictionary definitions to rare words so that the network can learn the best way to use them. We demonstrate improvements of up to 1.8 BLEU using bilingual dictionaries.
\end{abstract}

\section{Introduction}
\label{sec:intro}

Despite its successes, neural machine translation (NMT) still has unresolved problems. Among them is the problem of rare words, which are paradoxically very common because of Zipf's Law. In part, this is a problem intrinsic to data-driven machine translation because the system will inevitably encounter words not seen in the training data. In part, however, NMT systems seem particularly challenged by rare words, compared with older statistical models. 

One reason is that NMT systems have a fixed-size vocabulary, typically 10k--100k words; words outside this vocabulary are represented using a special symbol like \unk{}. Byte pair encoding (BPE) breaks rare words into smaller, more frequent subwords, at least allowing NMT to see them instead of \unk{} \citep{sennrich-etal-2016-neural}. But this by no means solves the problem; even with subwords, NMT seems to have difficulty learning translations of very rare words, possibly an instance of catastrophic forgetting \citep{mccloskey+cohen:1989}.

Humans deal with rare words by looking them up in a dictionary, and the idea of using dictionaries to assist machine translation is extremely old. From a statistical perspective, dictionaries are a useful complement to running text because the uniform distribution of dictionary headwords can smooth out the long-tailed distribution of running text. In pre-neural statistical machine translation systems, the typical way to incorporate bilingual dictionaries is simply to include them as parallel sentences in the training data. But (as we show), this does not work well for NMT systems.

We are aware of only a few previous attempts to find better ways to incorporate bilingual dictionaries in NMT. Some methods use dictionaries to synthesize new training examples \citep{zhang2016bridging,qi-etal-2018-pre,hamalainen+alnajjar:2019}. \citet{arthur-etal-2016-incorporating} extend the model to encourage it to generate translations from the (automatically extracted) dictionary. \citet{post+vilar:naacl2018} constrain the decoder to generate translations from the dictionary. What these approaches have in common is that they all treat dictionary definitions as target-language text, when, in fact, they often have properties very different from ordinary text. For example, CEDICT defines \zh{此致} (\emph{cǐzhì}) as ``(used at the end of a letter to introduce a polite salutation)'' which cannot be used as a translation. 

In this paper, we present an extension of the Transformer \citep{NIPS2017_7181} that ``attaches'' the dictionary definitions of rare words to their occurrences in source sentences. We introduce new position encodings to represent the nonlinear structure of a source sentence with its attachments. Then the unmodified translation model can learn how to make use of this attached information. We show that this additional information yields improvements in translation accuracy of up to 1.8 BLEU. 
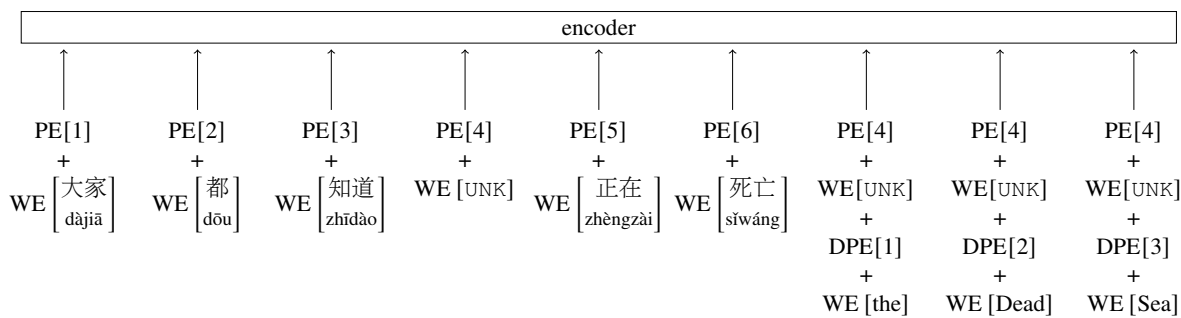
\begin{figure*}

    \newcommand{\zhen}[3]{\mbox{\begin{tabular}{@{}c@{}}\zh{#1} \\ \footnotesize #2 \end{tabular}}}

    \centering
    
    \scalebox{0.8}{%
    \begin{tikzpicture}[x=2.2cm]
    \tikzset{every node/.append style={anchor=north}}
    \node[draw,rectangle,minimum width=19cm,anchor=south] at (5,1.1) {encoder};
    \node (w1) at (1,0) {$\begin{array}{c}\textrm{PE}[1] \\ + \\ \textrm{WE}\left[\text{\zhen{大家}{dàjiā}{everyone}}\right]\end{array}$};
    \node (w2) at (2,0) {$\begin{array}{c}\textrm{PE}[2] \\ + \\ \textrm{WE}\left[\text{\zhen{都}{dōu}{all}}\right]\end{array}$};
    \node (w3) at (3,0) {$\begin{array}{c}\textrm{PE}[3] \\ + \\ \textrm{WE}\left[\text{\zhen{知道}{zhīdào}{knows}}\right]\end{array}$};
    \node (w4) at (4,0) {$\begin{array}{c}\textrm{PE}[4] \\ + \\ \textrm{WE}\left[\unk\right]\end{array}$};
    \node (w5) at (5,0) {$\begin{array}{c}\textrm{PE}[5] \\ + \\ \textrm{WE}\left[\text{\zhen{正在}{zhèngzài}{is}}\right]\end{array}$};
    \node (w6) at (6,0) {$\begin{array}{c}\textrm{PE}[6] \\ + \\ \textrm{WE}\left[\text{\zhen{死亡}{sǐwáng}{dying}}\right]\end{array}$};
    \node (w7) at (7,0) {$\begin{array}{c}\textrm{PE}[4] \\ + \\ \textrm{WE}[\unk] \\ + \\ \textrm{DPE}[1] \\ + \\  \textrm{WE}\left[\text{the}\right]\end{array}$};
    \node (w8) at (8,0) {$\begin{array}{c}\textrm{PE}[4] \\ + \\ \textrm{WE}[\unk] \\ + \\ \textrm{DPE}[2] \\ + \\ \textrm{WE}\left[\text{Dead}\right]\end{array}$};
    \node (w9) at (9,0) {$\begin{array}{c}\textrm{PE}[4] \\ + \\ \textrm{WE}[\unk] \\ + \\ \textrm{DPE}[3] \\ + \\ \textrm{WE}\left[\text{Sea}\right]\end{array}$};
    
    \begin{scope}[->]
    \draw (w1.north) to +(0,1);
    \draw (w2.north) to +(0,1);
    \draw (w3.north) to +(0,1);
    \draw (w4.north) to +(0,1);
    \draw (w5.north) to +(0,1);
    \draw (w6.north) to +(0,1);
    \draw (w7.north) to +(0,1);
    \draw (w8.north) to +(0,1);
    \draw (w9.north) to +(0,1);
    \end{scope}
    \end{tikzpicture}}
    
    \caption{Our method attaches dictionary definitions to rare words. Here, the source sentence is \zh{大家\ 都\ 知道\ 死海\ 正在\ 死亡} (\emph{dàjiā dōu zhīdào Sǐhǎi zhèngzài sǐwáng}, \emph{Everyone knows that the Dead Sea is dying}). $\textrm{WE}[f]$ is the embedding of word $f$, $\textrm{PE}[p]$ is the encoding of position $p$, and $\textrm{DPE}[q]$ is the encoding of position $q$ within a dictionary definition. The rare word \zh{死海} (\emph{Sǐhǎi}) is replaced with \unk{} and defined as \emph{the Dead Sea}. The words of the definition are encoded with both the position of the defined word (4) and their positions within the definition.}
    \label{modelalign}
\end{figure*}

\section{Methods}

Our method is built on top of the Transformer~\citep{NIPS2017_7181}. For each unknown source word with an entry in the dictionary, we attach the first 50 tokens of the definition (discarding the rest of the definition) to the source sentence. As described below, we encode the definition so as to differentiate it from the source sentence proper and to record which source word the definition is attached to. We leave the task of deciding whether and how to use the definition up to the translation model, which we use without any modifications.

\subsection{Position encodings}

To differentiate the attached definitions from the source sentence itself, we use special position encodings.

An ordinary word $f$ at position $p$ is encoded, as usual, as $\textrm{E}[f] = \textrm{WE}[f] + \textrm{PE}[p]$, where $\textrm{WE}$ is the word embedding and $\textrm{PE}$ is the usual sinusoidal position encoding \citep{NIPS2017_7181}.

Suppose that word $f$ at position $p$ has an attached definition. Then word $d$ at position $q$ of the definition is encoded as \[\textrm{E}[d] = \textrm{WE}[f] + \textrm{PE}[p] + \textrm{WE}[d] + \textrm{DPE}[q],\] where $\textrm{DPE}$ is a position encoding scheme different from $\textrm{PE}$. We experimented with several schemes for $\textrm{DPE}$; in the experiments below, we learned a different encoding for each position \citep{gehring+:icml2017}.

See Figure~\ref{modelalign} for an illustration of the encoding of an example source sentence. Note that once all words have received their position encodings, their order does not matter, as the Transformer encoder is order-independent.
\begin{figure*}

    \newcommand{\zhen}[3]{\mbox{\begin{tabular}{@{}c@{}}\zh{#1} \\ \footnotesize #2 \end{tabular}}}

    \centering
    
    \scalebox{0.75}{%
    \begin{tikzpicture}[x=2.0cm]
    \tikzset{every node/.append style={anchor=north}}
    \node[draw,rectangle,minimum width=19cm,anchor=south] at (5.5,1.1) {encoder};
    \node (w1) at (1,0) {$\begin{array}{c}\textrm{PE}[1] \\ + \\ \textrm{WE}\left[\text{\zhen{大家}{dàjiā}{everyone}}\right]\end{array}$};
    \node (w2) at (2,0) {$\begin{array}{c}\textrm{PE}[2] \\ + \\ \textrm{WE}\left[\text{\zhen{都}{dōu}{all}}\right]\end{array}$};
    \node (w3) at (3,0) {$\begin{array}{c}\textrm{PE}[3] \\ + \\ \textrm{WE}\left[\text{\zhen{知道}{zhīdào}{knows}}\right]\end{array}$};
    \node (w4) at (4,0) {$\begin{array}{c}\textrm{PE}[4] \\ + \\ \textrm{WE}\left[\text{\zhen{死@@}{sǐ}{dead}}\right]\end{array}$};
    \node (w5) at (5,0) {$\begin{array}{c}\textrm{PE}[5] \\ + \\
    \textrm{WE}\left[\text{\zhen{海}{hǎi}{sea}}\right]\end{array}$};
    \node (w6) at (6,0) {$\begin{array}{c}\textrm{PE}[6] \\ + \\
    \textrm{WE}\left[\text{\zhen{正在}{zhèngzài}{is}}\right]\end{array}$};
    \node (w7) at (7,0) {$\begin{array}{c}\textrm{PE}[7] \\ + \\ \textrm{WE}\left[\text{\zhen{死亡}{sǐwáng}{dying}}\right]\end{array}$};
    \node (w8) at (8,0) {$\begin{array}{c}\textrm{PE}[4] \\ + \\ \textrm{WE}[\unk] \\ + \\ \textrm{DPE}[1] \\ + \\  \textrm{WE}\left[\text{the}\right]\end{array}$};
    \node (w9) at (9,0) {$\begin{array}{c}\textrm{PE}[4] \\ + \\ \textrm{WE}[\unk] \\ + \\ \textrm{DPE}[2] \\ + \\ \textrm{WE}\left[\text{Dead}\right]\end{array}$};
    \node (w10) at (10,0) {$\begin{array}{c}\textrm{PE}[4] \\ + \\ \textrm{WE}[\unk] \\ + \\ \textrm{DPE}[3] \\ + \\ \textrm{WE}\left[\text{Sea}\right]\end{array}$};
    
    \begin{scope}[->]
    \draw (w1.north) to +(0,1);
    \draw (w2.north) to +(0,1);
    \draw (w3.north) to +(0,1);
    \draw (w4.north) to +(0,1);
    \draw (w5.north) to +(0,1);
    \draw (w6.north) to +(0,1);
    \draw (w7.north) to +(0,1);
    \draw (w8.north) to +(0,1);
    \draw (w9.north) to +(0,1);
    \draw (w10.north) to +(0,1);
    \end{scope}
    \end{tikzpicture}}
    
    \caption{After BPE is applied to the sentence only the rare word \zh{死海} is broken up into \zh{死@@} and \zh{海} thus we attach the definition to \zh{死@@} also at position 4.}
    \label{modelalign_bpe}
\end{figure*}
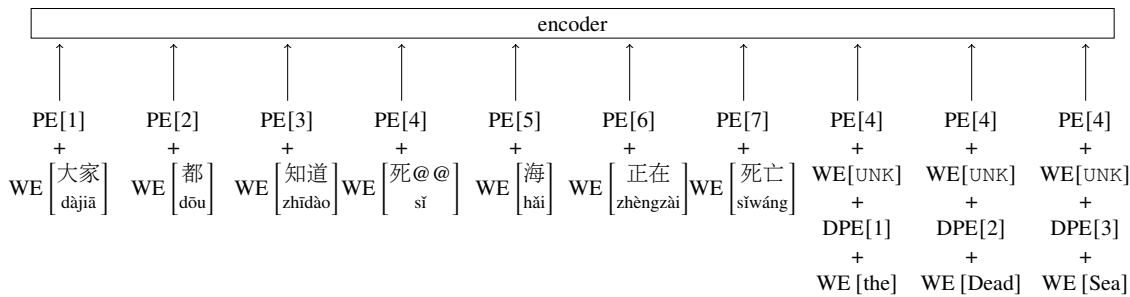

\begin{table*}
\centering \small
\begin{tabular}{ll|rrrr|rrr}
\toprule
& & \multicolumn{4}{c|}{lines} & \multicolumn{3}{c}{words} \\
Language & Task & train & dev & test & total & tokens & types & vocab \\
\midrule
Chi-Eng & Spoken  & 176,000 & 22,000 & 22,000 & 220k & 5.9M & 179k & 25k \\
& Science & 216,000 & 27,000 & 27,000 & 270k & 10.1M & 383k & 27k \\
& Laws & 176,000 & 22,000 & 22,000 & 220k & 17.4M & 98k & 22k \\
& News & 360,000 & 45,000 & 45,000 & 450k & 25.3M & 477k & 24k \\
& Education & 360,000 & 45,000 & 45,000 & 450k & 18.6M & 461k & 28k \\
& Subtitles & 240,000 & 30,000 & 30,000 & 300k & 6.6M & 147k & 27k \\
& Thesis & 240,000 & 30,000 & 30,000 & 300k & 17.2M & 613k & 27k \\
& UM-all & 1,993,500 & 221,500 & 5,000 & 2.2M & 101.3M & 1.3M & 33k \\
\midrule
Deu-Eng & Europarl-small & 160,000 & 20,000 & 20,000 & 200k & 10.9M & 151k & 16k \\
& Europarl-all & 1,440,000 & 180,000 & 197,758 & 1.8M & 98.6M & 475k & 16k \\
\bottomrule
\end{tabular}
\caption{Statistics of the various tasks we experimented on. Train/dev/test: number of lines selected for use as training, development, and test data (respectively). Toks: number of word tokens (source+target). Types: number of word types (source+target). Vocab: joint vocabulary size used in word-based experiments.}
\label{tab:data}
\end{table*}

\subsection{Subword segmentation}
\label{sec:bpemethod}

To apply our method to data that has been segmented using BPE, we face two new problems. First, since very few words are replaced with \unk, it is not sufficient only to attach definitions to \unk. How do we decide which words to attach definitions to? Second, if a word has been split into multiple subwords, the definition does not have a single attachment position. How do we represent the attachment position when encoding the definition?

To choose which words to define, we use a simple frequency threshold. If the frequency of a word is above the threshold, we do not attach any definitions. If it is at or below the threshold, we attach the definitions to the first subword. For example, in the sentence in Figure~\ref{modelalign_bpe}, only \zh{死海} (\emph{sǐhǎi}) is at or below the frequency threshold (here, 25), so we attach the definition of \zh{死海} to its first subword, \zh{死@@}.

\subsection{Fuzzy Matching}

In many languages, there are multiple morphologically inflected forms for each headword in the dictionary. Consequently, we extend our approach to find the closest possible dictionary entry for each rare word.

For each rare word, we try to find the dictionary headword with the lowest normalized Levenshtein distance to the rare word. The Levenshtein distance between two strings is the minimum number of insertions, deletions, or replacements needed to transform one string to the other; normalized Levenshtein distance divides the number of edits by the length of the longer string.\footnote{\url{https://pypi.org/project/textdistance/}} Thus, identical strings have a distance of 0, and completely different strings have a distance of 1.

When attaching the dictionary definitions, we multiply the DPEs by $(1-d)$, where $d$ is the normalized Levenshtein similarity. 

Since computing Levenshtein distance between the entire vocabulary and the entire dictionary would be prohibitively expensive, we used locality sensitive hashing \cite{10.5555/2787930} to approximate the search more efficiently.\footnote{\url{http://ekzhu.com/datasketch/lsh.html}} We convert the rare word into character trigrams, then into a vector using Minhash \cite{10.5555/2787930}. We then query for dictionary headwords using LSH with a Jaccard similarity score \cite{10.5555/2787930} of 0.5 or more.

\section{Experiments}

In this section, we describe our experiments on Chinese-English and German-English translation, comparing our methods -- \emph{Attach}, which uses exact matching, and \emph{Edit}, which uses fuzzy matching -- against two baselines. One baseline is the standard Transformer without any dictionary information (which we call \emph{Baseline}). The other baseline is the standard Transformer with the bilingual dictionaries included as parallel sentences in the training data (which we call \emph{Append}).

\subsection{Data: Chinese-English}

For Chinese-English, we used the UM-Corpus\footnote{\url{http://nlp2ct.cis.umac.mo/um-corpus/}} \citep{tian-etal-2014-um}, which has about 2M sentence pairs in eight different domains. Since rare words may be more frequent in certain domains, testing our model on different types of data may highlight the conditions where dictionaries can be helpful. We excluded the Microblog domain because of its length (only 5000 lines). For each of the other domains, we split the data into three parts: the first roughly 80\% for training (\emph{train}), the next 10\% for development (\emph{dev}), and the last 10\% for testing (\emph{test}). The task \emph{UM-all} combines all eight domains. The UM-Corpus provides a test set, which we used (\emph{test}), and we split the provided training data into two parts, the first 90\% for training (\emph{train}) and last 10\% for development (\emph{dev}). The exact line counts and other statistics are shown in Table~\ref{tab:data}.

We used the Stanford segmenter\footnote{\url{https://nlp.stanford.edu/software/segmenter.shtml}} \citep{stanfordsegmenter} for the Chinese data and the Moses tokenizer\footnote{\url{http://www.statmt.org/moses/}} for the English data. 

As a dictionary, we used CC-CEDICT,\footnote{\url{https://www.mdbg.net/chinese/dictionary?page=cedict}, downloaded 10/2018.} which has 116,493 entries. 
Each entry has a traditional Chinese headword (which we delete), a simplified Chinese headword, a pronunciation (which we delete), and one or more definitions. We process the definitions as follows:
\begin{itemize}
\item Remove substrings of the form \emph{abbr. for $c$}, where $c$ is a Chinese word.
\item If a definition contains \emph{see $c$} or \emph{see also $c$}, where $c$ is a Chinese word, replace it with the definition of $c$.
\item Remove everything in parentheses.
\item Remove duplicate definitions.
\item If the entry has no definitions left, delete the whole entry.
\item Concatenate all the definitions into a single string.
\end{itemize}
The resulting dictionary has 102,567 entries, each consisting of a Chinese headword and a single English definition.
We segmented/tokenized these in the same way as the parallel data.
The average definition length is five, and the maximum definition length is 107.

For example, consider the following CEDICT entries, where we have already removed traditional Chinese characters and pronunciations for clarity.
\begin{center}
\begin{tabular}{l>{\raggedright\arraybackslash}p{2.25in}}
\zh{三自} & /abbr. for \zh{三自爱国教会}, Three-Self Patriotic Movement/ \\
\zh{U盘} & /USB flash drive/see also \zh{闪存盘} \\
\zh{闪存盘} & /USB flash drive/jump drive/thumb drive/memory stick/
\end{tabular}
\end{center}
After cleaning, these would become
\begin{center}
\begin{tabular}{l>{\raggedright\arraybackslash}p{2.25in}}
\zh{三自} & Three-Self Patriotic Movement \\
\zh{U盘} & USB flash drive jump drive thumb drive memory stick \\
\zh{闪存盘} & USB flash drive jump drive thumb drive memory stick
\end{tabular}
\end{center}

\subsection{Data: German-English}

For German-English, we used the Europarl V7 dataset.\footnote{\url{http://statmt.org/europarl/}} We tokenized both sides of the data with the Moses tokenizer. Due to the size of the original Europarl dataset and the increased runtime from our method, we ran some experiments on only the first 200k lines of the dataset, denoted in result tables as \emph{Europarl-small}, while the full Europarl data is called \emph{Europarl-all}. We split both into three parts: the first roughly 80\% for training, the next 10\% for development, and the last 10\% for testing. Some statistics of the data are shown in Table~\ref{tab:data}.

We used the German-English dictionary from Stardict,\footnote{\url{http://download.huzheng.org/freedict.de/}} which is derived from Freedict\footnote{\url{https://freedict.org/}} and has 81,628 entries. In this dictionary, the headwords have notes in parentheses indicating things like selectional restrictions; we deleted all of these. Unlike with CEDICT, we did not delete any material in definitions, nor did we resolve cross-references, which were very rare. As before, we removed blank entries and merged multiple definitions into a single line. We tokenized both headwords and definitions with the Moses tokenizer. The final dictionary size is 80,737 entries, with an average definition length of 2.9 and a maximum definition length of 88. 

For example, the entry:
\begin{center}
\begin{tabular}{ll}
(Aktien) zusammenlegen & to merge (with)
\end{tabular}
\end{center}
would become
\begin{center}
\begin{tabular}{ll}
zusammenlegen & to merge (with)
\end{tabular}
\end{center}

\newcommand{\insignif}{\rlap{$^{=}$}}

\begin{table*}
\begin{center}
\begin{tabular}{l|rr|rr|rr}
\toprule
Task & \multicolumn{2}{c|}{Baseline} & \multicolumn{2}{c|}{Append} & \multicolumn{2}{c}{Attach}  \\
 & BLEU & MacroF1 & BLEU & MacroF1 & BLEU & MacroF1\\
\midrule
Spoken  & 14.0 & 13.0   & 13.0 & 12.0       & \textbf{14.6} & \textbf{13.7} \\
Science & 8.2 & 5.6     & 8.7 & 5.6\insignif  & \textbf{9.2} & \textbf{6.0} \\
Laws    & \textbf{30.3} & 11.5 & 28.0 & 10.6 & 29.6 & \textbf{11.8} \\
News    & 10.8 & 4.4    & 10.4 & 4.3        & \textbf{11.4} & \textbf{4.9} \\
Education & 8.8 & 5.7   & 8.8\insignif & 5.5   & \textbf{9.6} & \textbf{6.2} \\
Subtitles & 18.8 & 15.5 & 16.5 & 12.8       & \textbf{19.4} & \textbf{16.4}\\
Thesis & 10.0 & 4.3     & 9.7 & 4.1         & \textbf{10.5} & \textbf{4.5} \\
UM-all & 16.5 & 18.8    & 17.1 & 18.9\insignif & \textbf{17.3} & \textbf{19.9} \\
\midrule
Europarl-small & 28.3 & 15.8 & 28.0 & 15.4 & \textbf{29.0} & \textbf{16.8} \\
Europarl-all & 29.1 & 7.5 & 29.0 & 7.5\insignif & \textbf{30.0} & \textbf{8.0} \\
\bottomrule
\end{tabular}
\end{center}
\caption{Results on word-based translation. Our method (Attach) significantly improves over the baseline in all tasks other than Laws. By constrast, appending the dictionary to the parallel data (Append) performs worse in most tasks. Differences to the baselines are significant for all tasks except where marked with =. The highest BLEU score and the highest MacroF1 score in each row are written in boldface.}
\label{tab:wordbased}
\end{table*}

\subsection{Implementation and details}

We used Witwicky,\footnote{\url{https://github.com/tnq177/witwicky}} an open-source implementation of the Transformer, with all of its default hyperparameters. We use the same random seed in each experiment. We modified it to attach dictionary definitions as described above. The code and our cleaned dictionaries are available under an open-source license.\footnote{\url{https://github.com/xjz92/Attach_first_bpe}}

For BPE-based translation, we used joint BPE with 16k operations. For word-based translation, we set each system's vocabulary size close to the vocabulary size of the corresponding BPE-based system. For example, the Spoken dataset with 16k BPE applied to the training data has 25,168 word types, so we limited the word-based model to 25,000 word types. The vocabulary size we chose for each data set is shown in Table \ref{tab:data}.

For all tasks except UM-all and Europarl-all, we trained for 20 epochs, and used the model with the highest dev BLEU to translate the test set. Due to the massive increase in training data on the UM-all and Europarl-all datasets, we only trained for 10 epochs. Otherwise, the settings are the same across all experiments.

We report BLEU \citep{papineni-etal-2002-bleu} and MacroF1 \citep{gowda-etal-2021-macro} scores of detokenized outputs against raw references. MacroF1 is the F1 score between the number of correct translated word \emph{types} between a reference and test dataset, giving a clearer picture of a system's ability to translate rare words. Both scores are computed using \citeauthor{gowda-etal-2021-macro}'s fork of SacreBLEU.\footnote{\url{https://github.com/isi-nlp/sacrebleu}} We perform significance testing with bootstrap resampling using 1000 samples, with a significance level of~$0.05$.

\subsection{Results: Word-Based}
\label{sec:wordbased}

Table \ref{tab:wordbased} shows results on word-based translation. The \emph{Append} column shows that simply appending the bilingual dictionary to the parallel training data is unhelpful for all tasks, except UM-all. For UM-all, \emph{Append} does improve BLEU but not MacroF1. By contrast, our method improves accuracy significantly over \emph{Baseline} and \emph{Append} across all tasks except Laws. For Laws, our method improves MacroF1 but not BLEU.

\subsection{Results: BPE-Based}

\begin{table*}
\begin{center}
\begin{tabular}{l|rr|rr|rrr}
\toprule
Task & \multicolumn{2}{c|}{Baseline} & \multicolumn{2}{c|}{Append}& \multicolumn{3}{c}{Attach}  \\
 & BLEU & MacroF1 & BLEU & MacroF1 & BLEU & MacroF1 & freq \\
\midrule
Spoken & 17.3 & 15.3 & 15.4 & 13.1 & \textbf{17.6} & \textbf{17.8} & 25 \\
Science & 13.4 & 18.0 & 12.2 & 13.4 & \textbf{15.2} & \textbf{22.2} & 10 \\
Laws & 30.1 & 13.7 & 26.8 & 10.6 & \textbf{31.0} & \textbf{14.4} & 15 \\
News & 12.6 & 12.8 & 11.8 & 12.2 & \textbf{13.2} & \textbf{14.5} & 20 \\
Education & 13.3 & 10.7 & 12.3 & 9.5 & \textbf{13.9} & \textbf{12.6} & 25 \\
Subtitles & \textbf{21.6} & 17.3 & 18.5 & 13.5 & 21.5 & \textbf{18.5} & 15 \\
Thesis & 15.6 & 17.2 & 14.9 & 16.3 & \textbf{16.1} & \textbf{18.0} & 15 \\
UM-all & 20.9 & 22.8 & 20.8\insignif & 22.9\insignif & \textbf{21.1} & \textbf{23.6} & 20 \\
\bottomrule
\end{tabular}
\end{center}
\caption{Results on BPE-based Chinese-English translation. Our method (Attach) improves significantly over the baseline in all tasks except for Subtitles. Appending the dictionary to the parallel data (Append) performs worse. All differences are significant except when marked with =. The highest BLEU score and the highest MacroF1 score in each row are written in boldface.}
\label{tab:bpebased}
\end{table*}
\begin{table*}
\small
\begin{center}
\begin{tabular}{l|rr|rr|rrr|rrr}
\toprule
Task & \multicolumn{2}{c|}{Baseline} & \multicolumn{2}{c|}{Append} & \multicolumn{3}{c|}{Attach} & \multicolumn{3}{c}{Edit} \\
 & BLEU & MacroF1 & BLEU & MacroF1 & BLEU & MacroF1 & freq & BLEU & MacroF1 & freq \\
\midrule
Europarl-small & 33.5 & 25.9 & 31.7 & 23.1 & 33.6\insignif & 26.9 & 15 & \textbf{33.8} & \textbf{27.5} & 15 \\
Europarl-all & 36.4 & 25.5 & 36.3 & 25.0\insignif & 36.5 & 25.4\insignif & 15 & \textbf{36.8} & \textbf{26.1} & 15 \\
\bottomrule
\end{tabular}
\end{center}
\caption{Results on BPE-based German-English translation. Our method with fuzzy matching significantly improves over the baseline. Differences to the baseline are significant except where marked with =.}
\label{tab:edit_distance}
\end{table*}

As described in Section~\ref{sec:bpemethod}, we attach definitions only for words whose frequency falls below a threshold. We found that the optimal frequency thresholds vary on different datasets, and there was no direct correlation to corpus size. For each dataset, we trained models using thresholds of $k=5$, $10$, $15$, $20$, $25$, and $50$. We reported the test scores of the models that had the highest BLEU score on the development dataset.

As before, we compared against the two baselines (\emph{Baseline} and \emph{Append}).
On Chinese-English (Table~\ref{tab:bpebased}), we only tested our \emph{Attach} model since Chinese has essentially no morphological inflection.
Appending the dictionary to the parallel data did worse than baseline, significantly so on all tasks except UM-all. By contrast, our model improved over the baseline significantly across all tasks except Subtitles.

On German-English (Table~\ref{tab:edit_distance}), \emph{Append} did significantly worse on both datasets, whereas our \emph{Attach} significantly improved BLEU on the full dataset and MacroF1 on the small dataset. Added fuzzy matching (\emph{Edit}), however, improved both BLEU and MacroF1 significantly on both the smaller and larger datasets.

\begin{table*}
    \centering
    \small
    \begin{tabular}{l l}
    \toprule
         Source & 1. \zh{不\ 只\ 是\ 科学家们\ 对\ 对称性(\unk{}) 感\ 兴趣\ 。}\\
         & 2. \zh{我\ 哥哥\ 听说\ 我们\ 做\ 了\ 火药(\unk{}) 。}\\
         & 3. \zh{有些\ 登山者\ 经过\ 他\ 身旁\ ，\ 打量(\unk{}) 了\ 他\ 一\ 番\ }\\
         \midrule
         Definitions & 1. \zh{对称性}: symmetry \\
         & 2. \zh{火药}: gunpowder(\unk{})\\
         & 3. \zh{打量}: to size sb(\unk{}) up to look sb(\unk{}) up and down to take the measure of to suppose to reckon \\
         \midrule
         Reference & 1. But it's not just scientists who are interested in symmetry. \\
         & 2. Well, my brother heard that we had made gunpowder.\\
         & 3. Some climbers had come by and looked at him,\\
         \midrule
         Baseline & 1. not just the scientists are interested in the \unk{} \\
         & 2. My brother had heard that we had done a \unk{}.\\
         & 3. And some of the climbers passed him and \unk{} him.\\
         \midrule
         Append & 1. It's not just about scientists who are interested in \unk{}.\\
         & 2. My brother has heard that we've done a lot of work.\\
         & 3. And some of the \unk{} came over and over and over again,\\
         \midrule
         Attach & 1. not only scientists are interested in symmetry in symmetry.\\
         & 2. My brother heard that we had done gunpowder. \\
         & 3. Some climbers passed by him and looked at him,\\
    \bottomrule     
    \end{tabular}
    \caption{Examples from word-based systems on the UM-Spoken data. In the first and second examples, the unknown words \zh{对称性} (\emph{du\`{i}ch\`{e}nx\`{i}ng}) and \zh{火药} (\emph{huǒyào}) cannot be translated by the baseline, even with the dictionary in the parallel data (Append). Our model successfully incorporates the dictionary definition \emph{symmetry}, but not \emph{gunpowder}, because it is unknown. In the third example, the definition is not suitable as a direct translation of the unknown word \zh{打量} (\emph{dǎliàng}), but our model generates the word \emph{looked}, apparently by picking out the word \emph{look} from the definition and inflecting it correctly for the context.}
    \label{tab:spok_word_examples}
\end{table*}

\begin{table*}
    \centering
    \small
    \begin{tabular}{l l}
    \toprule
         BPE Source & 1. \zh{不\ 只\ 是\ 科学家们\ 对\ 对@@\ 称@@\ 性\ 感\ 兴趣 。}\\
         & 2. \zh{我\ 哥哥\ 听说\ 我们\ 做\ 了\ 火@@\ 药\ 。 }\\
         & 3. \zh{有些\ 登@@\ 山@@\ 者\ 经过\ 他\ 身@@\ 旁\ ，\ 打@@\ 量\ 了\ 他\ 一\ 番\ }\\
         \midrule
         Definitions & 1. \zh{对称性}: sym@@ metry \\
         & 2. \zh{火药}: gun@@ powder\\
         & 3. \zh{打量}: to size s@@ b up to look s@@ b up and down to take the measure of to suppose to reck@@ on\\
         \midrule
         Reference & 1. But it's not just scientists who are interested in symmetry. \\
         & 2. Well, my brother heard that we had made gunpowder.\\
         & 3. Some climbers had come by and looked at him,\\
         \midrule
         Baseline & 1. Not just scientists are interested in respect to sex.\\
         & 2. My brother has heard of the drugs we made. \\
         & 3.  Some climbers pass him by the side, and they took him over,\\
         \midrule
         Append & 1. Not only scientists are interested in the symmetry of sex.\\
         & 2. My brother told us that we had done a fire. \\
         & 3. Some of the climber passed his feet, and he took a second,\\  
         \midrule
         Attach & 1. is not just scientists are interested in symmetry.\\
         & 2. My brother heard that we had done a gunpowder. \\
         & 3. Some climbers passed by him and looked at him,\\
    \bottomrule     
    \end{tabular}
    \caption{Examples from BPE-based systems on the UM-Spoken data. In the first two examples, the baseline system, even with the dictionary in the parallel data (Append), tries to translate the pieces of unknown words separately and incorrectly (e.g., \emph{fire}, \emph{pills}, \emph{sex}). Our model is able to translate the first and third examples correctly as in Table~\ref{tab:spok_word_examples}, as well as the second example.}
    \label{tab:spok_bpe_examples}
\end{table*}

\section{Analysis}

To further examine how our methods improve translation, we looked at some examples in our UM-Spoken dev set, shown in Table \ref{tab:spok_word_examples} (word-based) and Table \ref{tab:spok_bpe_examples} (BPE). The (\unk{}) tag next to dictionary definitions indicates that the word is outside of the system's vocabulary.

In the first example, \zh{对称性} (\emph{du\`{i}ch\`{e}nx\`{i}ng}, symmetry) is unknown to the word-based systems. Adding the definition to the parallel training data (\emph{Append}) does not help word-based translation because the word remains unknown, whereas our model correctly generates the translation \emph{symmetry}. With BPE, the word is broken into three pieces, so that the Append system can correctly generate the word \emph{symmetry}. But the third character (\zh{性}, \emph{xìng}) can also mean ``sex,'' and together with the following character (\zh{性感}, \emph{xìnggaň}) can mean ``sexy.'' This explains why the Baseline and Append systems incorrectly adds the words \emph{of sex}.

In the second example, \zh{火药} (\emph{huǒyào}, gunpowder) is unknown, and the definition word \emph{gunpowder} is also unknown. So none of the systems are able to translate this word correctly (though arguably our system's generation of \unk{} is preferable). When we switch to BPE, our model generates the correct translation. The other systems fail because this word splits into two very common words, \zh{火} (\emph{h\v{u}o}, fire), and \zh{药} (\emph{y\`{a}o}, drug), which the system tries to translate separately.

The third example shows what happens when we have a long definition that contains useful information, but is not suitable as a direct translation of the unknown word \zh{打量} (\emph{dǎliàng}). Here we see that our attachment model generates the word \emph{looked}, apparently by picking out the word \emph{look} from the definition and inflecting it correctly for the context. No other models were able to generate a word with a similar meaning.

Please see Appendix~\ref{sec:viz} for visualizations of the encoder-decoder attention for these three examples.

\begin{table*}
    \centering
    \small
    
    \begin{tabular}{l|l}
    \toprule
         Source & 1. Ich hoffe , dass diese Auslassung(\unk{}) korrigiert werden kann .\\
         & 2. Wäre das nicht eine Alternativlösung(\unk{}) ?\\
         \midrule
         Definitions & 1. Auslassung: omission(\unk)\\
         & 2. Alternativlösung: alternative solution\\
         \midrule
         Reference  & 1. I hope that this omission can be corrected.\\
         & 2. Would this not be an alternative solution?\\
         \midrule
         Baseline & 1. I hope that these \unk{} can be corrected. \\
         & 2. Would this not be a \unk{}?\\
         \midrule
         Append & 1. I hope that this \unk{} can be corrected.\\
         & 2. Would this not be a \unk{}?\\
         \midrule
         Attach & 1. I hope that this \unk{} can be corrected.\\
         & 2. Would this not be an alternative solution?\\

    \bottomrule     
    \end{tabular}
    \caption{Examples from word-based systems run on the Europarl-small data. In the first example, the dictionary defines unknown word \emph{Auslassung} with another unknown word, \emph{omission}, so neither adding the dictionary to the parallel data (Append) nor our model (Attach) benefits. In the second example, adding the dictionary definition of \emph{Alternativlösung} to the parallel data does not help, but our model is able to incorporate it.}
    \label{tab:parl_word_examples}
\end{table*}

\begin{table*}
    \centering
    \small
    
    \begin{tabular}{l|l}
    \toprule
         BPE source & 1. Ich hoffe , dass diese Aus@@ l@@ assung korrigi@@ ert werden kann .\\
         & 2. W@@ äre das nicht eine Altern@@ ativ@@ lösung ?\\
         \midrule
         Definitions & 1. Auslassung: om@@ is@@ sion\\
         & 2. Alternativlösung: alternative solution\\
         \midrule
         Reference  & 1. I hope that this omission can be corrected.\\
         & 2. Would this not be an alternative solution?\\
         \midrule
         Baseline & 1. I hope that this approval can be corrected. \\ 
         & 2. Would this not be a alternative solution?\\
         \midrule
         Append & 1. I hope that this interpretation can be corrected.\\
         & 2. Would this not be a alternative solution?\\
         \midrule
         Attach & 1. I hope that this omission can be corrected.\\
         & 2. Would this not be an alternative solution?\\
    \bottomrule     
    \end{tabular}
    \caption{Examples from BPE-based systems run on the Europarl-small data. In the first example, unlike in Table~\ref{tab:parl_word_examples}, the unknown word \emph{Auslassung} is not replaced with \unk{} but is split into subwords, which the baseline system as well as the system with the dictionary in its parallel data (Append) translate incorrectly. Our model successfully uses the dictionary definition, \emph{omission}. In the second example, BPE enables all models to translate the compound \emph{Alternatvlösung} correctly.}
    \label{tab:parl_bpe_examples}
\end{table*}

We also looked at a few examples from the Europarl-small dev set, shown in Table~\ref{tab:parl_word_examples} and~\ref{tab:parl_bpe_examples}. In the first example, the definition \emph{omission} was out of vocabulary, so our model was not able to perform any better than the baselines. However, in the BPE systems, our model was able to properly translate \emph{Auslassung} to \emph{omission} while none of the other baseline systems was able to. 

\section{Discussion}

In Section~\ref{sec:intro}, we mentioned several other methods for using dictionaries in NMT, all of which treat dictionary definitions as target-language text. An alternative approach to handling rare words, which avoids dictionaries altogether, is to use word embeddings trained on large amounts of monolingual data, like fastText embeddings \citep{bojanowski2017enriching}. \Citet{qi-etal-2018-pre} find that fastText embeddings can improve NMT, but there is a sweet spot (likely between 5k and 200k lines) where they have the most impact. They also find that pre-trained embeddings are more effective when the source and target languages are similar. 

We, too, experimented with using fastText word embeddings in our NMT system, but have not seen any improvements over the baseline -- perhaps because our datasets are somewhat larger than those used by \citet{qi-etal-2018-pre}. We also experimented with using dictionaries to improve word embeddings and found that the present approach, which gives the model direct access to dictionary definitions, is far more effective.

The most significant limitation of our method is runtime: because it increases the length of the source sentences, training and decoding take 2--3 times longer. Another limitation is that the effectiveness of this method depends on the quality and coverage of the dictionaries. 

In the future, we plan to experiment with additional resources, like thesauruses, gazetteers, or bilingual dictionaries with a different target language. Second, from our examples, we see that our model is able to select a snippet of the definition and adapt it to the target context (for example, by inflecting words), but further analysis is required to understand how much the model is able to do this.

\section{Conclusion}

In this paper, we presented a simple yet effective way to incorporate dictionaries into a Transformer NMT system, by attaching definitions to source sentences to form a nonlinear structure that the Transformer can learn how to use. We showed that our method can beat baselines significantly, by up to 1.8 BLEU. We also analyzed our system's outputs and found that our model is learning to select and adapt parts of the definition, which it does not learn to do when the dictionary is simply appended to the training data. 

\section*{Acknowledgements}

This paper is based upon work supported in part by the Office of the Director of National Intelligence (ODNI), Intelligence Advanced Research Projects Activity (IARPA), via contract \#FA8650-17-C-9116. The views and conclusions contained herein are those of the authors and should not be interpreted as necessarily representing the official policies, either expressed or implied, of ODNI, IARPA, or the U.S. Government. The U.S. Government is authorized to reproduce and distribute reprints for governmental purposes notwithstanding any copyright annotation therein.

\bibliographystyle{acl_natbib}
\bibliography{dictionaries}

\appendix

\section{Attention Visualizations}
\label{sec:viz}

\begin{figure*}
    \centering
    \begin{tabular}{@{}c@{}c@{}}
    word-based & BPE \\
    \includegraphics[width=0.5\textwidth]{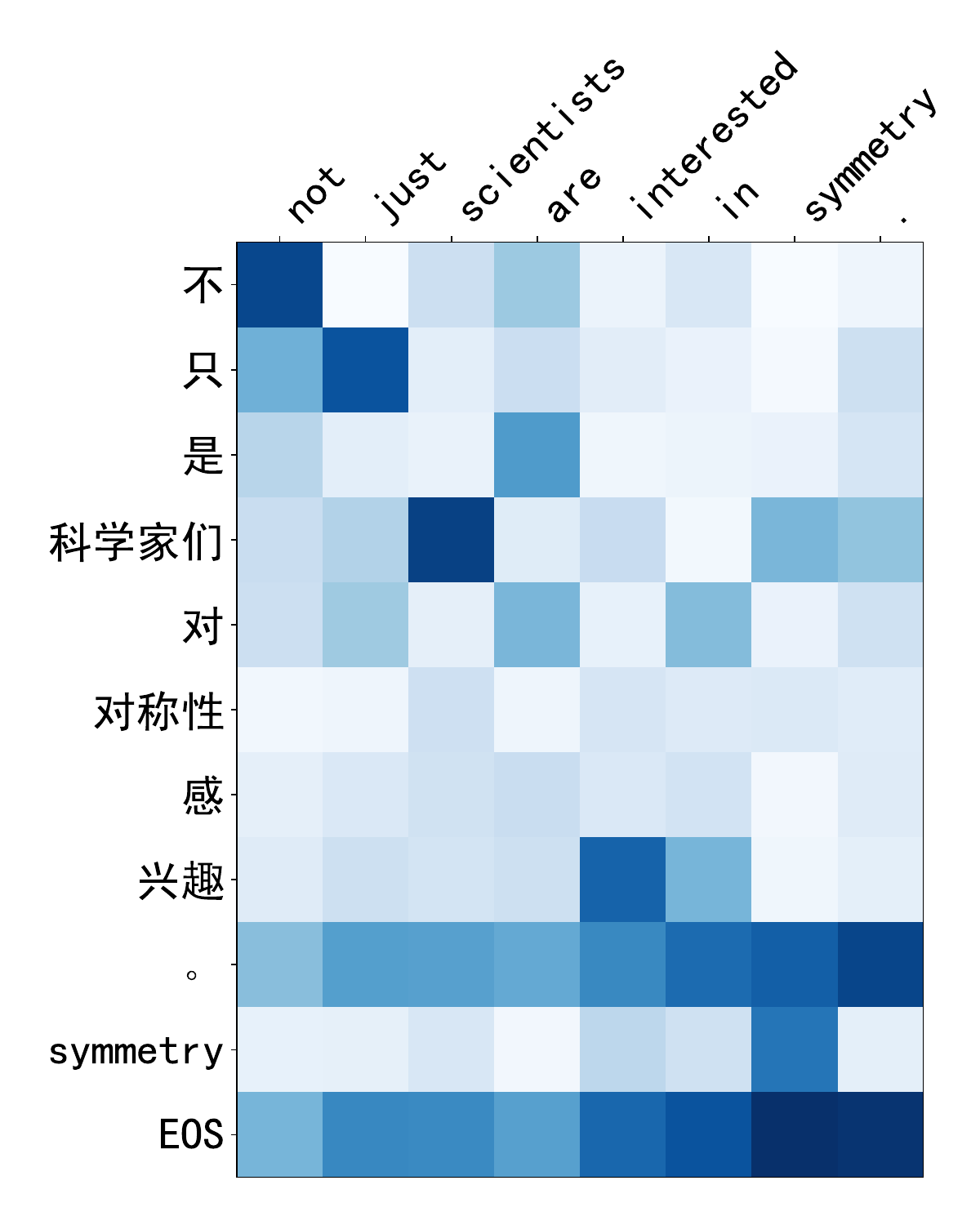} &     \includegraphics[width=0.5\textwidth]{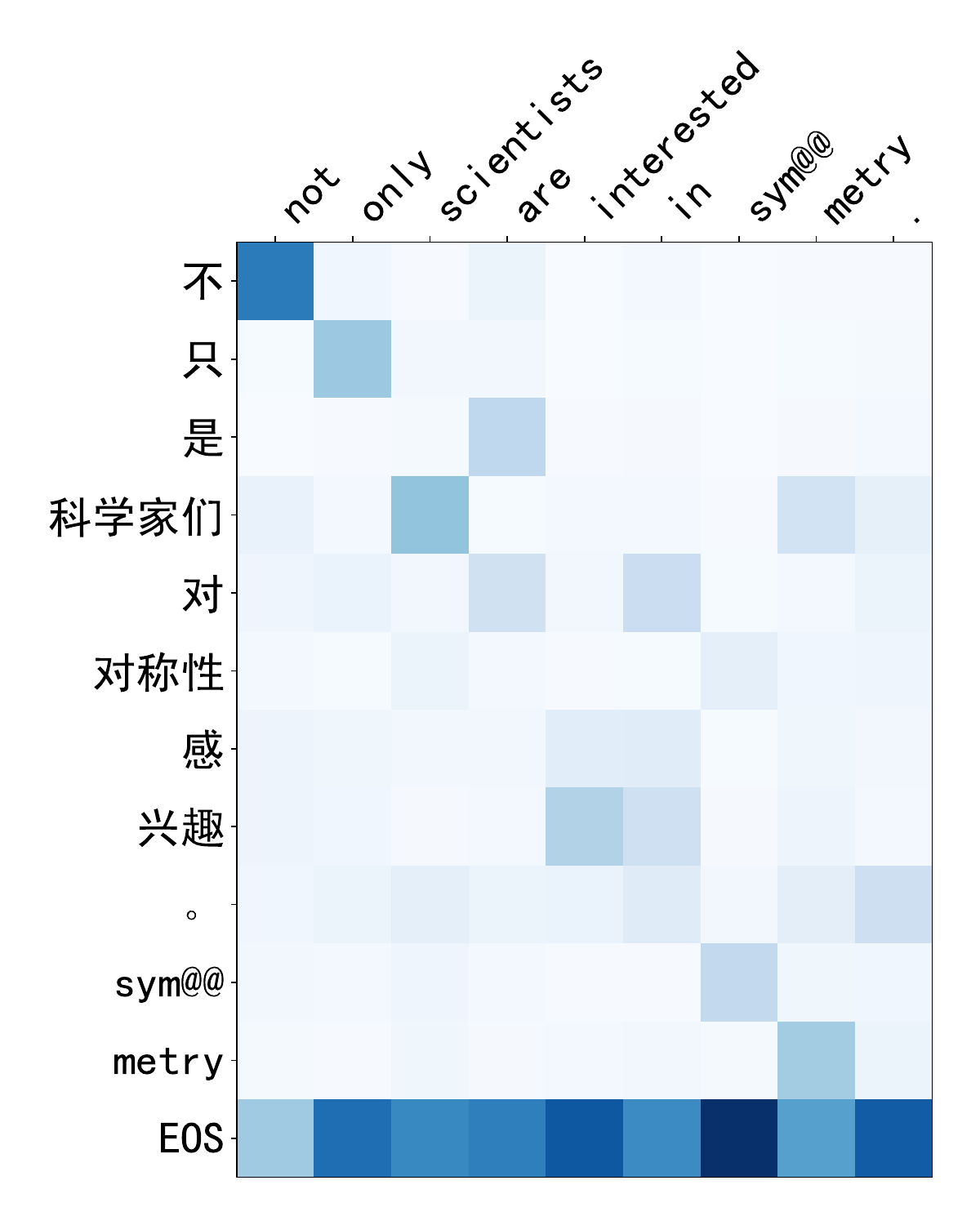} \\
    \includegraphics[width=0.5\textwidth]{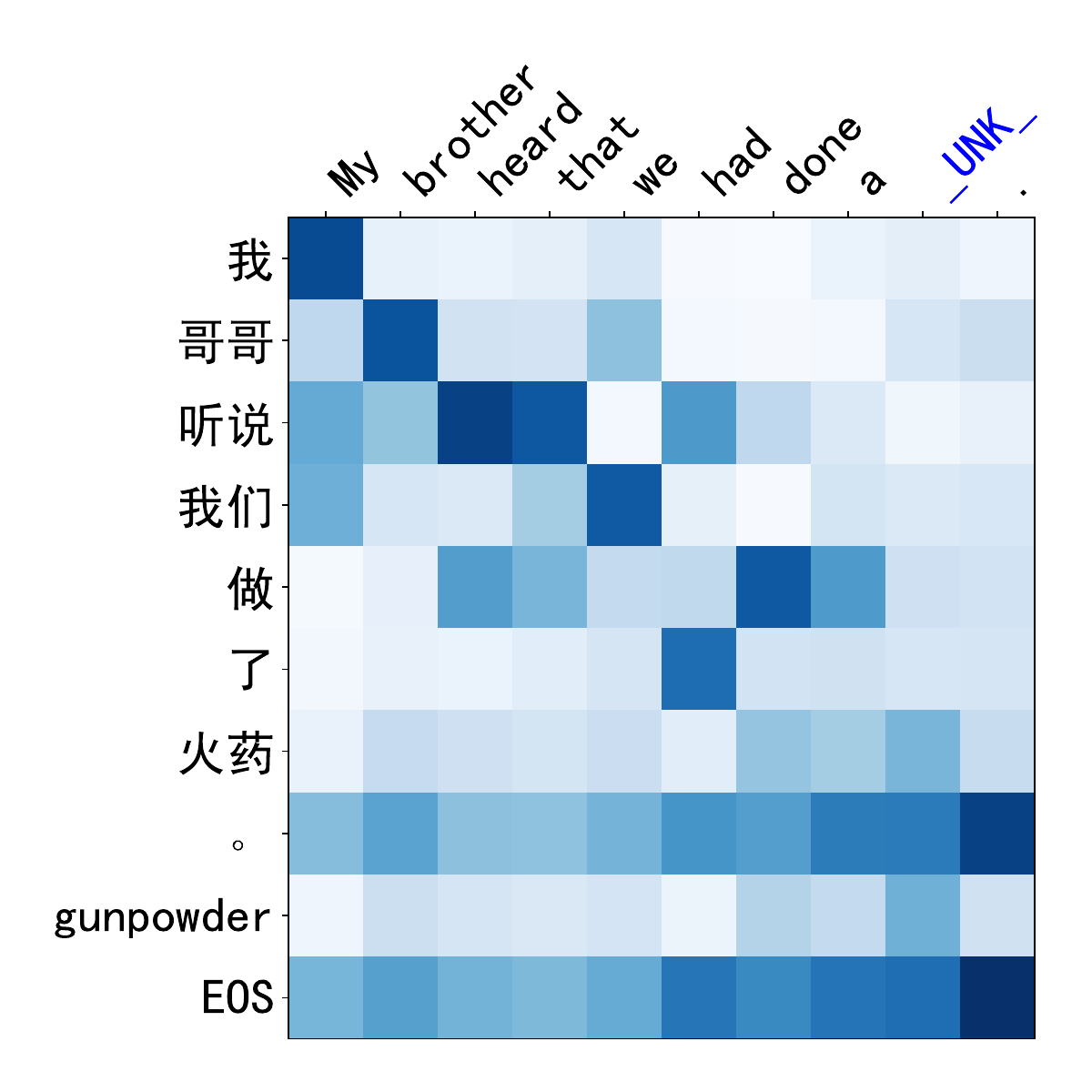} &        \includegraphics[width=0.5\textwidth]{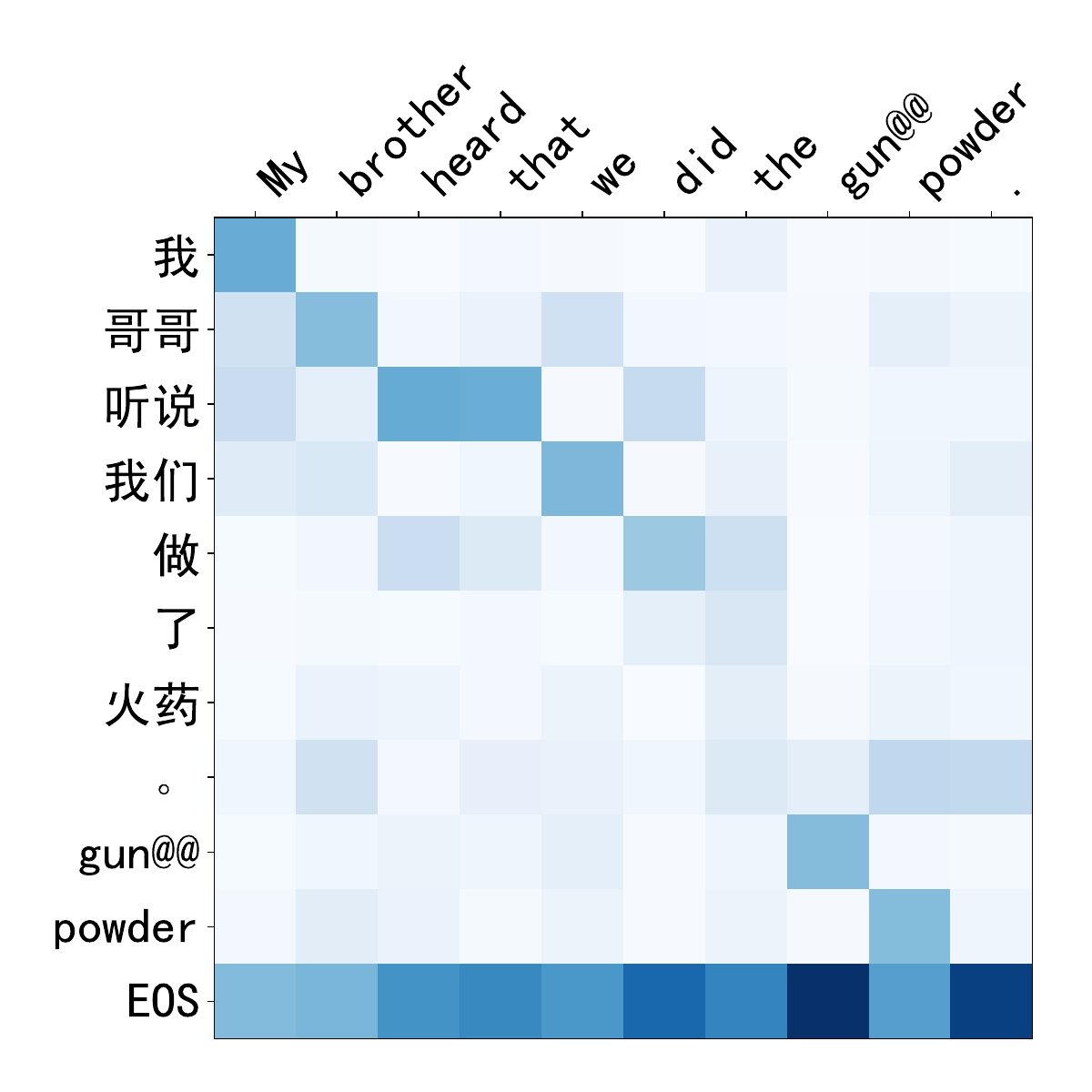} \\
    \end{tabular}
    \caption{Attention visualizations for the first two Chinese-English examples of Tables~\ref{tab:spok_word_examples} and \ref{tab:spok_bpe_examples}.}
    \label{fig:zhexamples1}
\end{figure*}

\begin{figure*}
    \centering
    \begin{tabular}{@{}c@{}c@{}}
    word-based & BPE \\
    \includegraphics[width=0.5\textwidth]{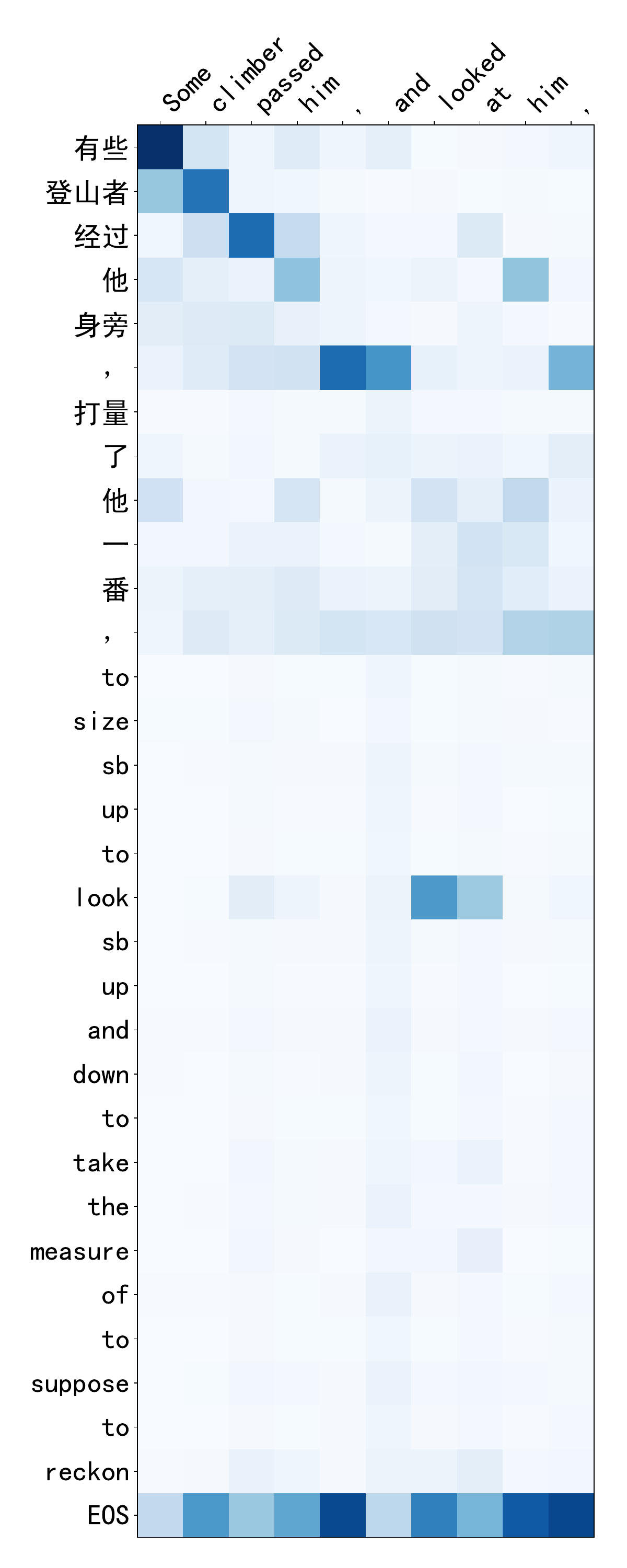} & 
    \includegraphics[width=0.5\textwidth]{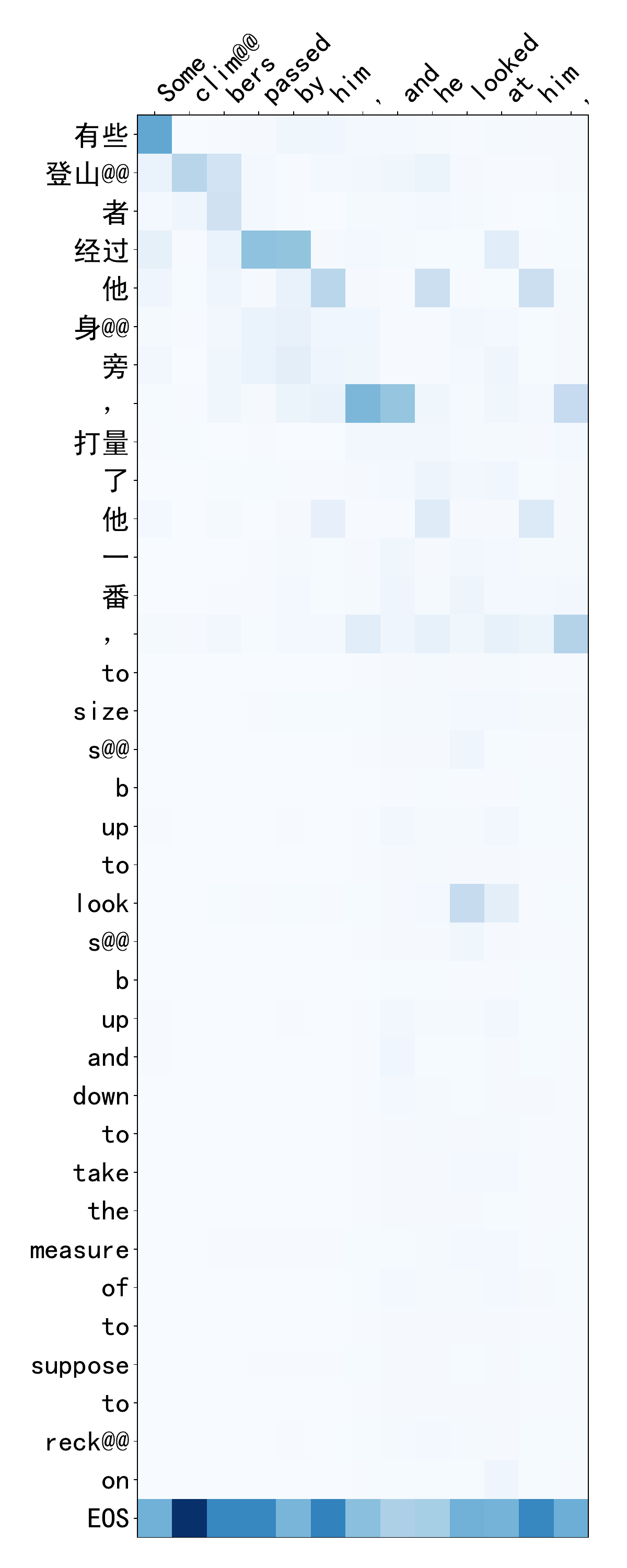} 
    \end{tabular}
    \caption{Attention visualizations for the third Chinese-English example of Tables~\ref{tab:spok_word_examples} and \ref{tab:spok_bpe_examples}.}
    \label{fig:zhexamples3}
\end{figure*}

Figures \ref{fig:zhexamples1} and \ref{fig:zhexamples3} show visualizations of the attention of our Attach model. They show the first layer of encoder-decoder attention when translating the three Chinese sentences of Tables~\ref{tab:spok_word_examples} and \ref{tab:spok_bpe_examples}. Note the translations are not exactly the same as shown above, because we used a beam size of one instead of the default of four.

\end{document}